\documentclass{article}
\usepackage{spconf,amsmath,graphicx,hyperref}
\usepackage{graphicx}
\usepackage{subcaption}
\usepackage{booktabs}
\usepackage{multirow}
\usepackage{algorithm}     
\usepackage{algorithmic}
\usepackage{amsmath}
\usepackage{amssymb}   
\usepackage{amsfonts}  
\usepackage{bm}        
\usepackage{placeins}
\usepackage[table]{xcolor}  

\title{ADVLA: Attention-Guided, Patch-Wise Sparse Adversarial Attacks on Vision-Language-Action Model}

\name{\centerline{Naifu Zhang$^{1}$, Wei Tao$^{2}$, Xi Xiao$^{1*}$, Qianpu Sun$^{1}$, Yuxin Zheng$^{1}$, Wentao Mo$^{1}$, Peiqiang Wang$^{1}$, Nan Zhang$^{3}$}}

\address{
$^{1}$Shenzhen International Graduate School, Tsinghua University, Shenzhen, China \\
$^{2}$Huazhong University of Science and Technology, Wuhan, China \\
$^{3}$Ping An Technology, Shenzhen, China\\
$^{*}$Corresponding author: xiaox@sz.tsinghua.edu.cn
}
\begin{document}
\ninept
\maketitle
\begin{abstract}
In recent years, Vision-Language-Action (VLA) models in the field of embodied intelligence have developed rapidly. Existing adversarial attack methods, however, require costly end-to-end training and often generate noticeable perturbation patches. To address these limitations, we propose \textbf{ADVLA}, a framework that directly applies adversarial perturbations on features projected from the visual encoder into the textual feature space. ADVLA efficiently disrupts downstream action predictions under low-amplitude constraints, and attention guidance allows the perturbations to be both focused and sparse. To achieve this, we introduce three strategies that enhance sensitivity, enforce sparsity, and concentrate perturbations. Experiments demonstrate that under an $L_\infty = 4/255$ constraint, ADVLA combined with Top-K masking modifies less than 10\% of the patches to achieve an attack success rate of nearly 100\% . The perturbations are focused on critical regions, remain almost imperceptible in the overall image, and a single-step iteration takes only about 0.06 seconds, significantly outperforming conventional patch-based attacks. In summary, ADVLA efficiently weakens the downstream action prediction of VLA models under low-amplitude and locally sparse conditions, avoiding the high training cost and conspicuous perturbations of traditional patch attacks, and demonstrates unique effectiveness and practical value for attacking VLA feature spaces.
\end{abstract}
\section{Introduction}
\label{sec:intro}
In recent years, the field of artificial intelligence has rapidly advanced, with large language models\cite{achiam2023gpt} \cite{touvron2023llama} (LLMs), an excellent text generation model, being among the most prominent directions. However, the real world contains multiple modalities, and with the development of LLMs, people are no longer satisfied with processing text alone, but instead hope that models can also handle visual information. Some studies achieve this by modality alignment or feature projection, allowing visual and textual information to be jointly modeled within a single model, endowing LLMs with the ability to "see." These models are referred to as vision-language models\cite{karamcheti2024prismatic} (VLMs).

On this basis, research on embodied intelligence has been inspired by vision-language models~\cite{cao2025efficient} (VLMs) and has developed vision-language-action models~\cite{kim2024openvla} \cite{liu2023libero} \cite{o2024open} (VLAs) capable of driving actions. For example, in robotic arm tasks, the system needs to process both natural language instructions and visual observations, and the VLA can aggregate this information at each time step and map it to action vectors, thereby guiding the robotic arm to complete tasks. The milestone work OpenVLA~\cite{kim2024openvla} adopts dual vision encoders with a large language model backbone to achieve end-to-end mapping in complex robotic tasks and achieves excellent performance. With the open-sourcing of OpenVLA~\cite{kim2024openvla}, more VLA models (such as OpenPI~\cite{black2024pi_0}, DexVLA~\cite{wen2025dexvla}, TinyVLA~\cite{wen2025tinyvla}) have emerged and are increasingly applied in real robotic scenarios.

Consequently, the safety issues of VLAs \cite{argus2025cvla} \cite{pertsch2025fast} have also become increasingly prominent. Unlike traditional vision models, VLAs map visual perception directly to action decisions. Small input deviations, such as natural noise, occlusions, environmental changes, or adversarial perturbations, may be amplified along the "perception-language alignment-action planning" chain, eventually leading to improper or even dangerous physical behaviors. In industrial, warehouse, human-robot collaboration \cite{zhao2023evaluating}, or medical assistance scenarios, this can result in mis-grasping, collisions, or even endanger safety.

When evaluating the safety of VLAs, adversarial attacks~\cite{joshi2021adversarial} \cite{yin2023vlattack} \cite{shayegani2023jailbreak} \cite{wu2024adversarial} \cite{zou2023universal}are a commonly used approach. By introducing small perturbations at the input, the model is forced to output incorrect predictions while maintaining high human-perceived similarity, thus revealing vulnerabilities. Although extensive studies exist for traditional vision models, such as pixel-level perturbations and adversarial patches, these mostly target classification or captioning tasks. Significant challenges remain for the vision-language-action chain in VLAs.

Recently, a few works~\cite{wang2024exploring} \cite{jones2025adversarial}have attempted to extend adversarial attacks to VLAs. One line of research focuses on image inputs, training an unconstrained visible adversarial patch end-to-end based on action outputs. However, such methods have obvious limitations: end-to-end optimization is time-consuming, the patches are visually conspicuous, and systematic exploration at the feature space level, crucial for perception, is lacking. Therefore, invisible attacks targeting the visual encoder feature space are still urgently needed.

\begin{figure*}[t]  
    \centering
    \includegraphics[width=\textwidth]{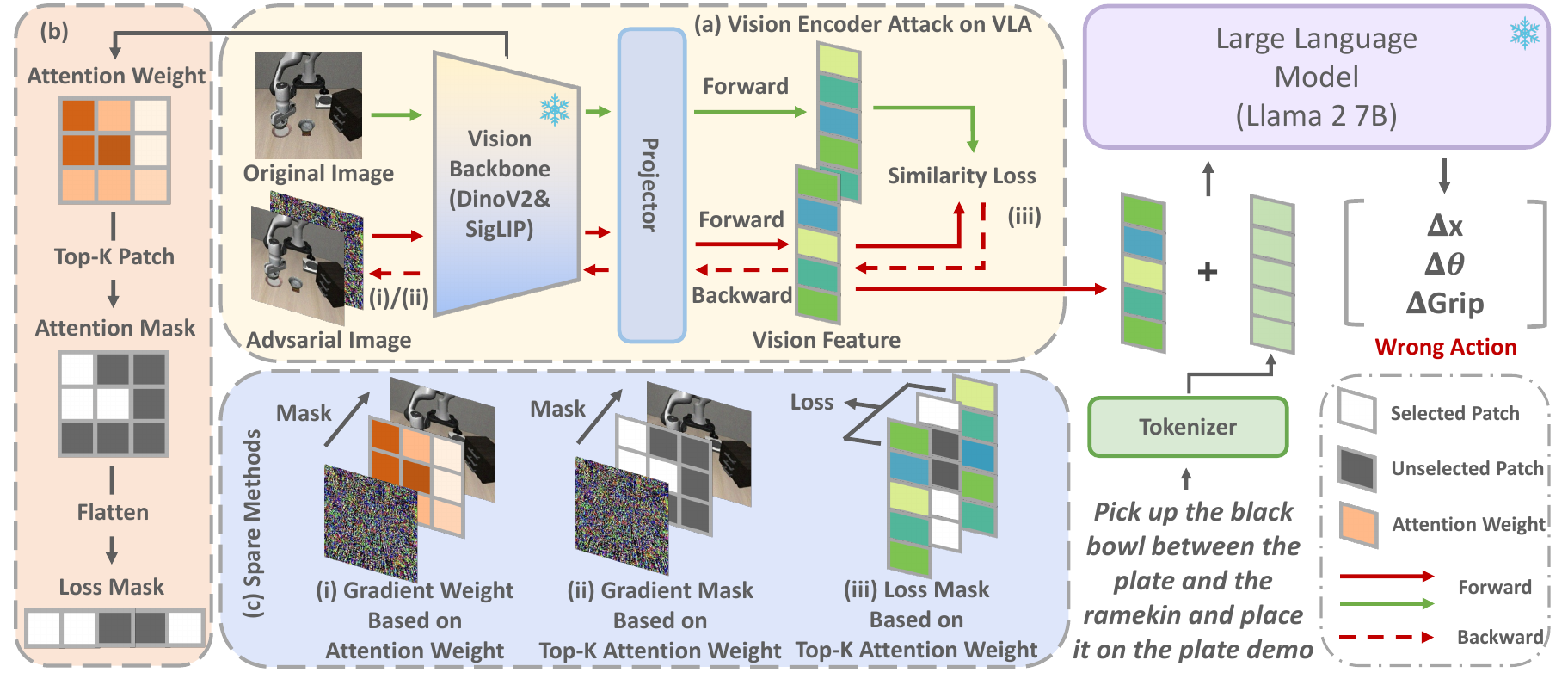}
    \caption{\textbf{The pipeline of ADVLA.} In Module (a), it is the main attack method of ADVLA, first, in every VLA step, we got original image from the environment and initialize random noise on it to create initial adversaial image, then we put original image in vision backbone and projector to generate clean vision feature, next, we iterate to put adversarial image in vision encoder to generate adversarial feature, and calculate similarity loss, so we can iterate perturbation based on Projection Gradient Descent (PGD) method and update adversarial image. In (b), we generate attention weight mask from vision backbone, which is composed of two Vision Transformers (ViT): DinoV2 \cite{oquab2023dinov2} and SigLIP\cite{zhai2023sigmoid}, openvla fused them to generate vision embeddings and project concated embddings to text feature space, we use one of the two ViTs to hook attention map and resize it to the size of image, then we set patchs which have top-k weights to 1 and others to 0, so we got attention mask, and then flatten it to get loss mask. (c), we show three methods, which use three mask on noise update and loss calculation respectively.}
    \vspace{-2mm}
    \label{fig:pipeline}
\end{figure*}

Overall, adversarial attacks on VLAs face three major challenges: 1) how to efficiently generate adversarial examples without long training; 2) how to achieve low-amplitude, imperceptible perturbations without reducing attack effectiveness; and 3) how to focus on model-sensitive regions to achieve sparse and efficient attacks. To this end, we propose ADVLA, a gray-box digital-domain adversarial attack framework that operates on the projected feature space of VLA visual encoders. Under the gray-box assumption, the attacker cannot access the entire model but can obtain the visual encoder parameters and gradients and apply perturbations to the input image, making the attack more realistic. ADVLA directly applies perturbations on the feature representations and combines PGD to achieve efficient attacks. Within the ADVLA framework, we implement three attention-guided strategies: first, gradient-weighted updates based on attention weights to focus perturbations on regions of model attention; second, sparse Top-K mask updates to limit the perturbation range and reduce visibility; third, loss computation only on key patch features to focus on sensitive areas. Each strategy can be used independently, flexibly meeting different attack needs. ADVLA achieves high success rates under low-amplitude perturbations, and each iteration takes approximately 0.06 seconds, significantly outperforming traditional patch-based methods.

The main contributions of this paper are as follows:  
\begin{itemize}
    \item We propose the ADVLA framework for gray-box attacks on VLAs in the visual feature space.
    \item We design three independent attention-guided strategies to achieve efficient, sparse, and imperceptible attacks.
    \item ADVLA achieves high success rates under strict constraints with less time while maintaining imperceptibility of perturbations.
\end{itemize}

\section{Method}
\label{sec:method}
The method of \textbf{ADVLA}, the adversarial attack framework implemented in the feature space of the VLA visual encoder. The whole pipline is presented as Fig.\ref{fig:pipeline}. Below we first present the attack assumptions and objectives in §\ref{sec:2.1}, then introduce the basic optimization framework of ADVLA in §\ref{2.2} and three attention-enhancement strategies in §\ref{sec:2.3} - §\ref{sec:2.5}. Finally, we provide the algorithm~\ref{alg:ADVLA} summarizes ADVLA with optional attention-guided tricks. 

\subsection{Threat Model and Attack Objective}
\label{sec:2.1}
This work constructs adversarial attacks against vision--language--action (VLA) models (represented by OpenVLA\cite{kim2024openvla}) under a gray-box setting: the attacker has access to the visual encoder and its projection layer (structure and parameters), and can read intermediate projected features, attention maps and obtain corresponding gradients; however, the attacker cannot access or modify internal parameters or gradients of the downstream LLM and action head, nor can they interact with the real physical environment. Attacks are performed in the digital domain, with image perturbations constrained by pixel amplitude (e.g., $L_\infty \le 4/255$), and optionally the proportion of modified patches is restricted to simulate sparse stealthy attacks.

Our objective is a feature-space untargeted attack: by minimizing the similarity between the clean image and the adversarial image in the visual$\to$text projection feature space, we induce a shift in the fused vision--language representation, thereby disrupting subsequent action prediction. The baseline method and the three attention-guided strategies are given in detail in Sec.~2.2 and Sec.~2.3--2.5 below.
\subsection{Vision Encoder Attack on OpenVLA}
\label{2.2}
At each control timestep, the model receives an input image $I$ captured by the camera. 
The image is first encoded by the vision backbone (DINOv2 and SigLIP in the case of OpenVLA), producing the unprojected visual embedding:  
\begin{equation}
    E = f_{\mathrm{vision}}(I).
\end{equation}

The embedding is then mapped into the text-aligned feature space via a projection layer:  
\begin{equation}
    F_{\mathrm{clean}} = g(E),
\end{equation}
where $F_{\mathrm{clean}}$ serves as the reference feature for this timestep and remains fixed during adversarial optimization.  

We adopt Projected Gradient Descent\cite{madry2017towards} (PGD) to apply perturbations on the input image. The perturbation is constrained by an $\ell_\infty$ bound $\delta_{\max}=k/255$, with the initial perturbation $\delta_0$ sampled from a Gaussian distribution and clipped to the allowed range. The update rule at the $t$-th iteration is:  
\begin{equation}
    \delta_t = \alpha \cdot \operatorname{sign}\!\big(\nabla_{I_t}\,\mathcal{L}(F_t, F_{\mathrm{clean}})\big),
\end{equation}
where
\begin{equation}
\begin{aligned}
F_t &= g\big(f_{\mathrm{vision}}(I_t)\big),\\
\mathcal{L}(F_t, F_{\mathrm{clean}}) &= 1 - \frac{F_t^\top F_{\mathrm{clean}}}{\|F_t\|\,\|F_{\mathrm{clean}}\| + \epsilon}.
\end{aligned}
\end{equation}

The updated perturbation is clipped to satisfy the constraint:  
\begin{equation}
    \delta_{t+1} = \operatorname{Clip}_{[-\delta_{\max},\,\delta_{\max}]}(\delta_t),
\end{equation}
and the adversarial image is obtained as:  
\begin{equation}
    I_{t+1}^{adv} = \operatorname{Clip}_{[0,1]}(I_t^{adv} + \delta_{t+1}).
\end{equation}

This process is conducted independently at each timestep: the reference feature $F_{\mathrm{clean}}$ is first computed and fixed for the current frame, then a number of PGD inner iterations are performed to generate the adversarial image $I^{\mathrm{adv}}$. The adversarial image is subsequently fed into the downstream modules, while in the next timestep a new frame is collected and the procedure is repeated.

\subsection{ADVLA-AW: Attention-Weighted Gradient (AW)}
\label{sec:2.3}
Let $G_t=\nabla_{I_t}\mathcal{L}$ denote the gradient on the image at iteration $t$.
We obtain the model attention map $A$ from the ViT (original size e.g. $[1,1,256]$), reshape it to patch-grid (e.g. $16\times16$) and resize to image resolution $(H\times W)$ by bicubic interpolation; denote this resized map as $\widetilde{A}\in\mathbb{R}^{H\times W}$. 
The attention-weighted gradient is:
\begin{equation}
    G_t^{\mathrm{AW}} = G_t \odot \widetilde{A}
\end{equation}
where $\odot$ denotes element-wise multiplication (with broadcasting over channels).
This emphasizes model-attended regions when updating the input image.

\subsection{ADVLA-TKM: Top-K Masked Gradient (TKM)}
\label{sec:2.4}

We select the top-$K$ patches according to the attention scores and convert them to a binary spatial mask $M_{\mathrm{topk}}\in\{0,1\}^{H\times W}$ (upsampled to image resolution). 
The masked gradient is then
\begin{equation}
    G_t^{\mathrm{TKM}} = G_t \odot M_{\mathrm{topk}}
\end{equation}
This restricts perturbation updates to the top-$K$ patches, achieving sparsity and lower visual conspicuity.

\subsection{ADVLA-TKL: Top-K Loss (TKL)}
\label{sec:2.5}

Instead of masking gradients, we mask features and compute the loss only on top-$K$ patch features. Let $E_t\in \mathbb{R}^{C\times N}$ be the projected visual features at timestep $t$ (with $N$ patches) and $F_{\mathrm{clean}}$ the clean reference features. Define a flattened binary patch-wise mask $M_{\mathrm{topk-flat}}\in\{0,1\}^N$ (broadcasted across channels) resized and flattened from $M_{\mathrm{topk_flat}}\in\{0,1\}^N$ . The Top-K loss is:
\begin{equation}
    \mathcal{L}_t^{\mathrm{TKL}} \;=\; 1 - \mathrm{sim}\!\big(F_t \odot M_{\mathrm{topk-flat}},\; F_{\mathrm{clean}} \odot M_{\mathrm{topk-flat}}\big)
\end{equation}
where $\mathrm{sim}(\cdot,\cdot)$ denotes cosine similarity. Optimizing this loss focuses the feature divergence on the selected patches.

\begin{algorithm}[t]
\caption{ADVLA: Attention-Guided Adversarial Attack on OpenVLA}
\label{alg:ADVLA}
\begin{algorithmic}[1]
\STATE \textbf{Input:} Clean image $\mathbf{I}$, step size $\alpha$, iterations $T$, top-$k$ ratio $k$
\STATE Initialize adversarial image: $\mathbf{I}_0 = \mathbf{I} + \text{UniformNoise}$
\FOR{$t = 0$ to $T-1$}
    \STATE Extract patch embeddings: $\mathbf{E}_t = f_\text{vision}(\mathbf{I}_t)$
    \STATE Project to text space: $\mathbf{F}_t = g(\mathbf{E}_t)$
    \STATE Compute similarity-based loss: $\mathcal{L}_t = 1 - \text{sim}(\mathbf{F}_t, \mathbf{F}_\text{clean})$
    \STATE Compute gradient: $\mathbf{G}_t = \nabla_{\mathbf{I}_t} \mathcal{L}_t$
    \STATE \textbf{Optionally apply attention-guided trick:}
    \IF{Use ADVLA-AW} 
        \STATE $\mathbf{G}_t \leftarrow \mathbf{G}_t^\text{AW}$
    \ENDIF
    \IF{Use ADVLA-TKM} 
        \STATE $\mathbf{G}_t \leftarrow \mathbf{G}_t^\text{TKM}$
    \ENDIF
    \IF{Use ADVLA-TKL} 
        \STATE $\mathcal{L}_t \leftarrow \mathcal{L}_t^\text{TKL}$
        \STATE Recompute $\mathbf{G}_t = \nabla_{\mathbf{I}_t} \mathcal{L}_t$
    \ENDIF
    \STATE Update adversarial image: $\mathbf{I}_{t+1} = \mathbf{I}_t + \alpha \cdot \text{sign}(\mathbf{G}_t)$
    \STATE Project $\mathbf{I}_{t+1}$ back to valid range
\ENDFOR
\STATE \textbf{Output:} Adversarial image $\mathbf{I}_T$
\end{algorithmic}
\end{algorithm}

\section{Experiments}
\label{sec:experiments}

\subsection{Experimental Setup}
\textbf{Datasets.} We conduct evaluations on the LIBERO dataset. LIBERO is a simulation benchmark covering four task suites (Spatial, Object, Goal, and Long), each consisting of 10 tasks. For each task, 50 rollouts are executed, yielding a total of 500 trials per suite.  

\noindent\textbf{Victim Models.} We select the publicly available and representative OpenVLA models as targets, including four variants independently trained on the four LIBERO suites.  

\noindent\textbf{Simulation Environment.} All experiments are performed in the standard simulation environment provided by LIBERO.  

\noindent\textbf{Evaluation Metrics.} For task execution evaluation, we set the maximum number of steps in the training data as the timeout condition. Following the LIBERO benchmark, we adopt Success Rate (SR) and report Failure Rate (FR = 1 - SR) as the primary metric.  

\noindent\textbf{Hardware.} All experiments are conducted on a single NVIDIA H100 GPU; the peak memory usage for attacking a single sample (including model loading) is approximately 17\,GB.

\begin{table}[t]
\centering
\caption{Evaluation of adversarial attacks on OpenVLA across different Libero benchmarks. 
We report Failure Rate (FR, ↑). Higher FR indicates stronger attack effectiveness. Results highlighted in gray denote the best performance.}
\label{tab:main_results}
\resizebox{\linewidth}{!}{%
\begin{tabular}{lccccc}
\toprule
Method & Spatial & Object & Goal & Long & Avg. \\
\midrule
Clean         & 15.30\% & 11.60\% & 20.80\% & 46.30\% & 23.50\% \\
Random Noise  & 17.20\% & 11.80\% & 22.40\% & 49.20\% & 25.15\% \\
\rowcolor{gray!20}
UADA\cite{wang2024exploring}  & 100.0\% & 100.0\% & 100.0\% & 100.0\% & 100.0\% \\
\midrule
\rowcolor{gray!20}
ADVLA        & 100.0\% & 100.0\% & 100.0\% & 100.0\% & 100.0\% \\
\rowcolor{gray!20}
ADVLA-AW     & 100.0\% & 100.0\% & 100.0\% & 100.0\% & 100.0\% \\
\rowcolor{gray!20}
ADVLA-TKM    & 100.00\%  & 100.00\%  & 100.00\%  & 100.00\%  & 100.00\% \\
ADVLA-TKL    & 99.40\%  & 99.40\%  & 99.60\%  & 100.00\%  & 99.60\% \\
\bottomrule
\end{tabular}%
}
\end{table}

\begin{table}[t]
\caption{Comparison of attack performance under different noise onstraints $\epsilon$ settings. Random noise apply uniform noise.Victim models are different models in Libero. The metric is FR $\uparrow$.}
\label{table2}
\centering
\resizebox{\linewidth}{!}{
\begin{tabular}{l l ccccc}
\toprule
$\epsilon$ & Method & Spatial & Object & Goal & Long & Avg. \\
\midrule
\multirow{5}{*}{2/255} 
& Random Noise & 17.80\% & 11.40\% & 21.00\% & 46.80\% & 24.25\% \\
& ADVLA & 30.40\% & 30.20\% & 36.00\% & 60.00\% & 39.15\% \\
& ADVLA-AW & 39.60\% & 38.80\% & 45.00\% & 66.00\% & 47.35\% \\
& ADVLA-TKM & 30.40\% & 30.00\% & 37.00\% & 59.00\% & 39.10\% \\
& ADVLA-TKL & 29.80\% & 29.80\% & 35.00\% & 58.00\% & 38.15\% \\
\midrule
\multirow{5}{*}{4/255} 
& Random Noise & 17.20\% & 11.80\% & 22.40\% & 49.20\% & 25.15\% \\
& ADVLA & 87.00\% &86.80\% &90.00\% & 99.00\% & 90.70\% \\
& ADVLA-AW & 89.80\% & 89.00\% & 91.80\% & 99.00\% & 92.40\% \\
& ADVLA-TKM & 88.60\% & 87.60\% & 90.80\% & 98.50\% & 91.38\% \\
& ADVLA-TKL & 81.40\% & 80.80\% & 91.00\% & 96.90\% & 87.53\% \\
\midrule
\multirow{5}{*}{8/255}
& Random Noise & 14.40\% & 11.60\% & 21.00\% & 48.70\% & 23.93\% \\
& ADVLA & 100.00\% & 100.00\% & 100.00\% & 100.00\% & 100.00\% \\
& ADVLA-AW & 100.00\% & 100.00\% & 100.00\% & 100.00\% & 100.00\% \\
& ADVLA-TKM & 100.00\% & 100.00\% & 100.00\% & 100.00\% & 100.00\% \\
& ADVLA-TKL & 100.00\% & 100.00\% & 100.00\% & 100.00\% & 100.00\% \\
\bottomrule
\end{tabular}}
\end{table}

\begin{table}[t]
\caption{Comparison of attack performance under constraint $\epsilon$ = 4/255 with different iteration settings. The step size is 1/255 and the metric is FR $\uparrow$.}
\label{table3}
\centering
\resizebox{\linewidth}{!}{
\begin{tabular}{l l ccccc}
\toprule
iter & Method & Spatial & Object & Goal & Long & Avg. \\
\midrule
& Random Noise & 17.80\% & 11.40\% & 21.00\% & 46.80\% & 24.25\% \\
\midrule
\multirow{4}{*}{4} 
& ADVLA & 87.00\% & 86.80\% & 90.00\% & 99.00\% & 90.70\% \\
& ADVLA-AW & 89.80\% & 89.00\% & 91.80\% & 99.0\% & 92.40\% \\
& ADVLA-TKM & 88.60\% & 87.60\% & 90.80\% & 98.50\% & 91.38\% \\
& ADVLA-TKL & 81.30\% & 80.80\% & 91.0\% & 97.0\% & 87.53\% \\
\midrule
\multirow{4}{*}{5} 
& ADVLA & 99.40\% &99.60\% &99.40\% & 100.00\% & 99.75\% \\
& ADVLA-AW & 100.00\% & 100.00\% & 100.00\% & 100.00\% & 100.00\% \\
& ADVLA-TKM & 99.40\% &99.60\% & 99.40\% & 100.00\% & 99.60\% \\
& ADVLA-TKL & 99.40\% & 99.80\% & 99.60\% & 100.00\% & 99.70\% \\
\midrule
\multirow{4}{*}{6}
& ADVLA & 100.00\% & 100.00\% & 100.00\% & 100.00\% & 100.00\% \\
& ADVLA-AW & 100.00\% & 100.00\% & 100.00\% & 100.00\% & 100.00\% \\
& ADVLA-TKM & 100.00\% & 100.00\% & 100.00\% & 100.00\% & 100.00\% \\
& ADVLA-TKL & 99.40\% & 99.40\% & 99.60\% & 100.00\% & 99.60\% \\
\bottomrule
\end{tabular}}
\end{table}

\subsection{Main Results}

\textbf{Quantitative Results.} Under the setting of maximum perturbation bound $\epsilon = 4/255$ and 6 iterations, ADVLA and its three extensions effectively perturb visual features and significantly degrade downstream action predictions, leading to substantially increased failure rates. As shown in Table~\ref{tab:main_results}, in some suites, the success rate drops to nearly 0\% and the average failure rate approaches 100\%, showing performance comparable to the UADA method.  
 
\subsection{Analysis}
\noindent\textbf{Sensitivity Analysis.} Table \ref{table2} reports results under different perturbation constraints. For fair comparison, the perturbation bound is set to $k/255$, with the number of iterations $k$ and step size $\alpha = 1/255$. The results show that failure rates increase consistently with larger $\epsilon$, at $\epsilon = 8/255$, models on multiple suites nearly collapse. Table \ref{table3} further analyzes the impact of iteration numbers under $\epsilon = 4/255$. Increasing iterations enhances the attack success, but good performance can already be achieved with fewer than six steps.  

\begin{figure}[!t]
    \centering
    \includegraphics[width=\columnwidth]{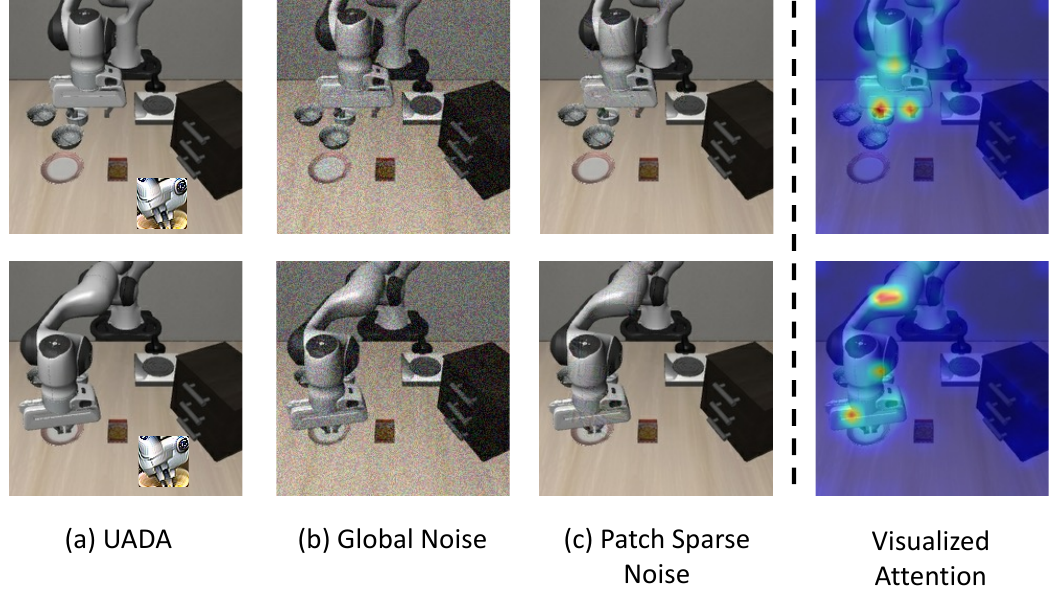}
    \caption{\textbf{Image samples in experiments.} For visualization purposes, all perturbation images are amplified for clarity (for display only). The column (a):  UADA\cite{wang2024exploring}, which apply an unlimited trained patch on observation images, and it's very easy to be found. The column (b): ADVLA with fully noise, it's easy to see there is noise on it. The column (c): ADVLA-TKM, we select top-10\% patches based on attention weight, as we see, the noise is almost invisible, just noise on top-k patches. Last column is the visualized attention weight of vision-backbone in vla, as shown in pictures, ViTs always mainly focus on the robot arms.}
    \label{fig:sample}
\end{figure}

\noindent\textbf{Time Analysis} The average time per inner-loop iteration of ADVLA is approximately 0.06 seconds, enabling fast adversarial evaluation. In contrast, UADA requires about 15 hours on a single H100 (batch size = 16) to generate one patch for the LIBERO-Spatial suite, making it infeasible for real-time robustness assessment. ADVLA thus achieves several orders of magnitude acceleration. 

\noindent\textbf{Case Study.} Fig.\ref{fig:sample} illustrates the visualizations of different perturbation strategies (global perturbation, sparse patch perturbation, and UADA patches). For visualization purposes, perturbations are amplified. The results indicate that UADA patches are highly noticeable, global perturbations remain perceptible, whereas sparse patch perturbations are almost imperceptible while still causing strong attack effects. This demonstrates that ADVLA achieves both stealthiness and effectiveness. 
\section{Conclusion}
\label{sec:conclusion}
We proposed \textbf{ADVLA}, an attention-guided adversarial attack framework for vision-language-action (VLA) models that perturbs visual features in projection space. By integrating gradient weighting, sparse Top-K masking, and focused loss masking, ADVLA achieves efficient, sparse, and visually imperceptible perturbations. Experiments on the LIBERO benchmark show that ADVLA nearly collapses model performance under strict perturbation constraints while being faster and stealthier than traditional patch attacks, highlighting the urgent need for stronger defenses in embodied intelligence.

\section*{Ethical Statement}
\label{sec:ethical}
This work studies adversarial attacks on vision-language-action (VLA) models with the goal of revealing vulnerabilities and promoting robustness. 
All experiments were conducted in simulation environments, without deployment on physical robots that could endanger human safety. 
The proposed methods are intended solely for research in trustworthy AI and should not be used for malicious purposes in any real-world deployment.

\bibliographystyle{IEEEbib}
\bibliography{strings,refs}

\end{document}